\documentclass[11pt,a4paper]{article}
\usepackage[hyperref]{acl2020}
\usepackage{times}
\usepackage{latexsym}

\usepackage{url}
\usepackage{graphicx}
\usepackage{amsmath,amssymb,amsfonts}
\usepackage{mathtools}
\usepackage{url}
\usepackage{siunitx}
\usepackage{textcomp}
\usepackage{booktabs}
\usepackage{graphicx}
\usepackage{caption}
\usepackage{footnote}
\usepackage{xcolor}
\usepackage{xspace}
\usepackage{float}
\usepackage{placeins}
\usepackage{svg}
\usepackage{tabularx}
\usepackage{multirow}
\usepackage[symbol]{footmisc}

\usepackage{microtype}

\aclfinalcopy % Uncomment this line for the final submission
\definecolor{hilite}{RGB}{0, 100, 200}
\newcommand{\bfblue}[1]{{\bf \textcolor{hilite}{#1}}}

\title{Two Birds, One Stone: A Simple, Unified Model for Text Generation\\ from Structured and Unstructured Data}
  
\author{Hamidreza Shahidi,$^{1,2}$ Ming Li,$^{1,2}$ \and Jimmy Lin$^{1,2}$   \vspace{0.2cm}\\
$^{1}$ David R. Cheriton School of Computer Science, University of Waterloo\\
$^{2}$ RSVP.ai\\
{\tt \{h24shahi, mli, jimmylin\}@uwaterloo.ca}}

\date{}

\begin{document}
\maketitle
\begin{abstract}
A number of researchers have recently questioned the necessity of increasingly complex neural network (NN) architectures. In particular, several recent papers have shown that simpler, properly tuned models are at least competitive across several NLP tasks.
In this work, we show that this is also the case for text generation from structured and unstructured data.
We consider neural table-to-text generation and neural question generation (NQG) tasks for text generation from structured and unstructured data, respectively.
Table-to-text generation aims to generate a description based on a given table, and NQG is the task of generating a question from a given passage where the generated question can be answered by a certain sub-span of the passage using NN models. 
Experimental results demonstrate that a basic attention-based seq2seq model trained with the exponential moving average technique achieves the state of the art in {\it both} tasks.
Code is available at \url{https://github.com/h-shahidi/2birds-gen}.
\end{abstract}

\section{Introduction}
Recent NLP literature can be characterized as increasingly complex neural network architectures that eke out progressively smaller gains over previous models.
Following a previous line of research \cite{melis2018on,mohammed2017strong,adhikari2019rethinking}, we investigate the necessity of such complicated neural architectures. In this work, our focus is on text generation from structured and unstructured data by considering description generation from a table and question generation from a passage and a target answer.

\begin{figure}[t!]
\centering{\includegraphics[scale=0.55]{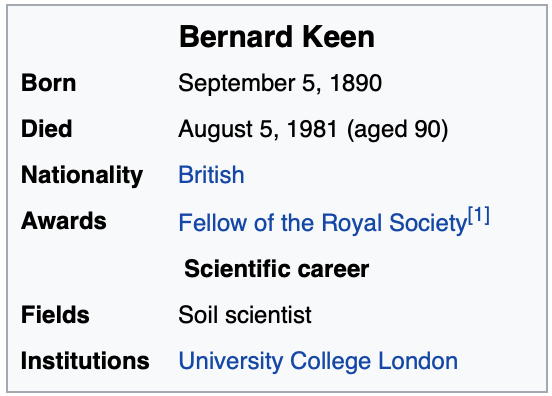}}\\
\vspace{0.1cm}
\raggedright
{\bf \small Target Output:} \\
{\small Sir Bernard Augustus Keen FRS (5~September 1890 \textendash{} 5~August 1981) was a British soil scientist and Fellow of University College London.}
\caption{An example infobox from the \textsc{WikiBio} dataset and the corresponding target output description.}
\label{fig:1}
\end{figure}

More specifically, the goal of the neural table-to-text generation task is to generate biographies based on Wikipedia infoboxes (structured data). An infobox is a factual table with a number of fields (e.g., name, nationality, and occupation) describing a person. For this task, we use the \mbox{\textsc{WikiBio}} dataset \cite{lebret-etal-2016-neural} as the benchmark dataset.
Figure \ref{fig:1} shows an example of a biographic infobox as well as the target output textual description.

\begin{table}[t!]
\centering
\resizebox{0.45\textwidth}{!}{
\begin{tabularx}{\columnwidth}{X}
\hline
\textbf{Passage:} Hydrogen is commonly used in power stations \color{blue} as a coolant in  generators \color{black} due to a number of favorable properties that are a direct result of its light diatomic molecules.  \\
\hline
\textbf{Answer:} as a coolant in generators \\
\hline
\textbf{Question:} How is hydrogen used at power stations? \\
\hline
\end{tabularx}
}
\caption{A sample (passage, answer, question) triple from the SQuAD dataset.}
\label{table:1}
\vspace{-0.5cm}
\end{table}

Automatic question generation aims to generate a syntactically correct, semantically meaningful and relevant question from a natural language text and a target answer within it (unstructured data). This is a crucial yet challenging task in NLP that has received growing attention due to its application in improving question answering systems \cite{duan2017question,tang2017question,tang2018learning}, providing material for educational purposes \cite{heilman2010good}, and helping conversational systems to start and continue a conversation \cite{mostafazadeh-etal-2016-generating}. We adopt the widely used SQuAD dataset \cite{rajpurkar-etal-2016-squad} for this task.
Table \ref{table:1} presents a sample (passage, answer, question) triple from this dataset.

Prior work has made remarkable progress on both of these tasks. However, the proposed models utilize complex neural architectures to capture necessary information from the input(s).
In this paper, we question the need for such sophisticated NN models for text generation from inputs comprising structured and unstructured data.

Specifically, we adopt a bi-directional, attention-based seq2seq model \cite{DBLP:journals/corr/BahdanauCB14} equipped with a copy mechanism \cite{DBLP:conf/acl/GuLLL16} {\it for both tasks}. We demonstrate that this model, together with the exponential moving average (EMA) technique, achieves the state of the art in both neural table-to-text generation and NQG.
Interestingly, our model is able to achieve this result even without using any linguistic features.

Our contributions are two-fold:
First, we propose a {\it unified} NN model for text generation from structured and unstructured data 
and show that training this model with the EMA technique leads to the state of the art in neural table-to-text generation as well as NQG.
Second, because our model is, in essence, the primary building block of previous models, our results show that some previous papers propose needless complexity, and that gains from these previous complex neural architectures are quite modest.
In other words, the state of the art is achieved by careful tuning of simple and well-engineered models, not necessarily by adding more complexity to the model, echoing the sentiments of~\citet{Lipton:1807.03341v2:2018}. 

\section{Related Work}\label{related_work}

In this section, we first discuss previous work for neural table-to-text generation and then NQG.

\subsection{Neural Table-to-Text Generation}

Recently, there have been a number of end-to-end trainable NN models for table-to-text generation.
\citet{lebret-etal-2016-neural} propose an n-gram statistical language model that incorporates field and position embeddings to represent the structure of a table. However, their model is not effective enough to capture long-range contextual dependencies while generating a description for the table.

To address this issue, \citet{liu2018table} suggest a structure-aware seq2seq model with local and global addressing on the table. 
While local addressing is realized by content encoding of the model's encoder and word-level attention, global addressing is accomplished by field encoding using a field-gating LSTM and field-level attention.
The field-gating mechanism incorporates field information when updating the cell memory of the LSTM units.

\citet{Liu_Luo_Xia_Ma_Chang_Sui_2019} utilize a two-level hierarchical encoder with coarse-to-fine attention to model the field-value structure of a table. They also propose three joint tasks (sequence labeling, text auto-encoding, and multi-label classification) as auxiliary supervision to capture accurate semantic representations of the tables.

In this paper, similar to \citet{lebret-etal-2016-neural}, we use both content and field information to represent a table by concatenating the field and position embeddings with the word embedding. 
Unlike \citet{liu2018table}, we don't separate local and global addressing by using specific modules for each, but rather adopt the EMA technique and let the bi-directional model accomplish this implicitly, exploiting the natural advantages of the model.

\subsection{Neural Question Generation}

Previous NQG models can be classified into rule-based and neural-network-based approaches. \citet{du-etal-2017-learning} propose a seq2seq model that is able to achieve better results than previous rule-based systems without taking the target answer into consideration.
\citet{zhou2017neural} concatenate answer position indicators with the word embeddings to make the model aware of the target answer.
They also use lexical features (e.g., POS and NER tags) to enrich their model's encoder. In addition, \citet{song2018leveraging} suggest using a multi-perspective context matching algorithm to further leverage information from explicit interactions between the passage and the target answer.

More recently, \citet{DBLP:conf/aaai/KimLSJ19} use answer-separated seq2seq, which replaces the target answer in the passage with a unique token to avoid using the answer words in the generated question.
They also make use of a module called keyword-net to extract critical information from the target answer.
Similarly, \citet{liu2019learning} propose using a clue word predictor by adopting graph convolution networks to highlight the imperative aspects of the input passage.

Our model is architecturally more similar to \citet{zhou2017neural}, but with the following distinctions:\
(1) we do not use additional lexical features, (2) we utilize the EMA technique during training and use the averaged weights for evaluation, (3) we do not make use of the introduced maxout hidden layer, and (4) we adopt LSTM units instead of GRU units.
These distinctions, along with some hyperparameter differences, notably the optimizer and learning rate, have a considerable impact on the experimental results (see Section \ref{results}).

\begin{figure}[t!]
\centering
\includegraphics[scale=0.62]{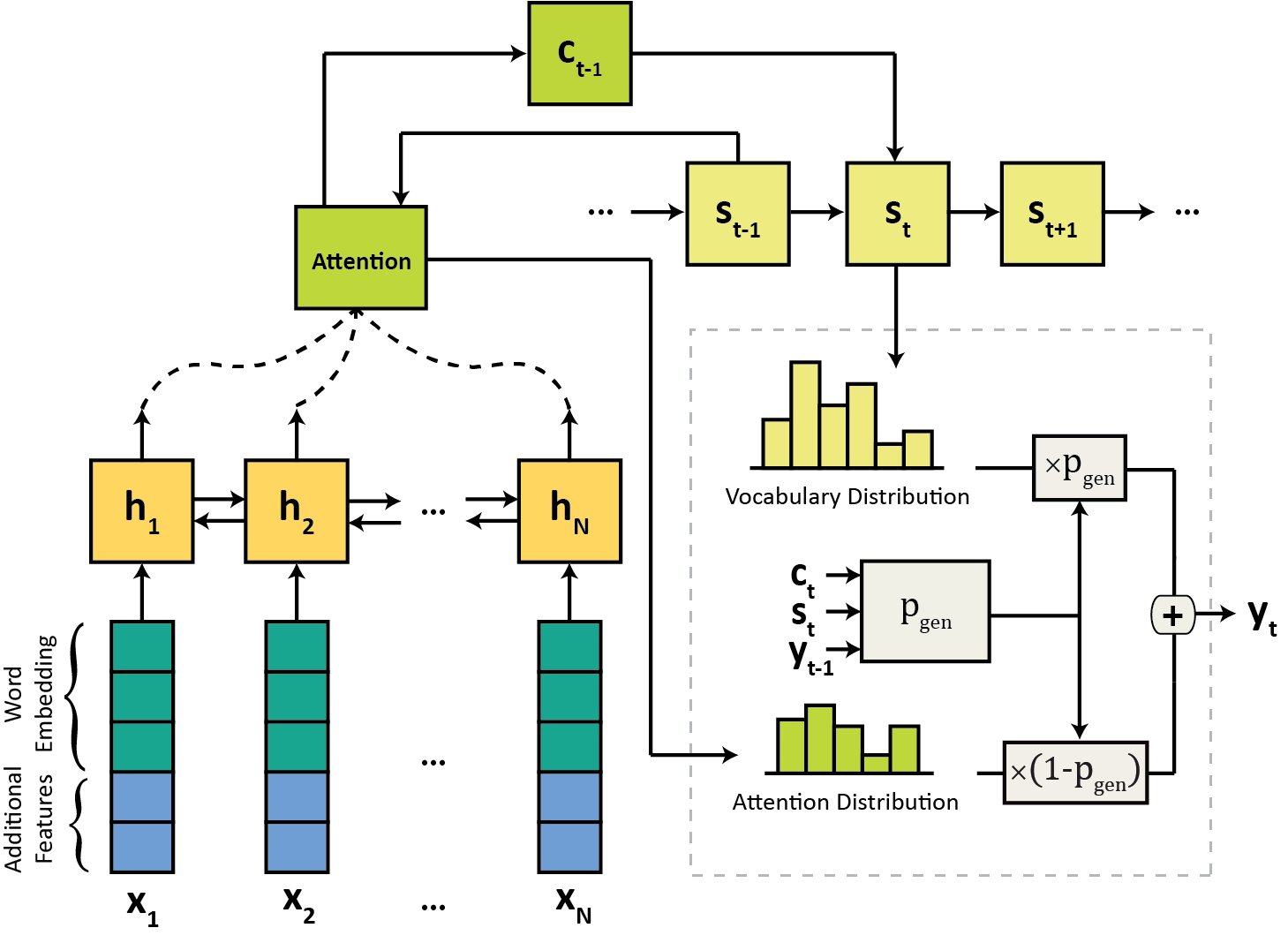}
\caption{An overview of our model.}
\label{fig:2}
%\vspace{-.4cm}
\end{figure}

\section{Model: Seq2Seq with Attention and a Copy Mechanism}

In this section, we introduce a simple but effective attention-based seq2seq model for both neural table-to-text generation and NQG. Figure \ref{fig:2} provides an overview of our model.

\subsection{Encoder}

Our encoder is a bi-directional LSTM (BiLSTM) whose input $x_t$ at time step $t$ is the concatenation of the current word embedding $e_t$ with some additional task-specific features.

For neural table-to-text generation, additional features are field name $f_t$ and position information $p_t$, following \citet{lebret-etal-2016-neural}.
The position information itself is the concatenation of $p^+_t$, which is the position of the current word in its field when counting from the left, and $p^-_t$, when counting from the right.
Considering the word \textit{University}, in Figure \ref{fig:1}, as an example, it is the first word from the left and the third word from the right in the \textit{Institutions} field.
Hence, the structural information of this word would be \{\textit{Institutions}, 1, 3\}.
Thus, the input to the encoder at time step $t$ for this task is $x_t = [e_t;f_t;p^+_t;p^-_t]$, where $[.;.]$ denotes concatenation along the feature dimension.

For NQG, similar to \citet{zhou2017neural}, we use a single bit $b_t$, indicating whether the t\textsuperscript{th} word in the passage belongs to the target answer, as an additional feature.
Hence, the input at time step $t$ is $x_t = [e_t;b_t]$.
Remarkably, unlike previous work~\cite{song2018leveraging,DBLP:conf/aaai/KimLSJ19}, we do not use a separate encoder for the target answer to have a {\it unified} model for both tasks.

\subsection{Attention-Based Decoder}

Our decoder is an attention-based LSTM model \cite{DBLP:journals/corr/BahdanauCB14}. Due to the considerable overlap between input and output words, we use a copy mechanism \cite{DBLP:conf/acl/GuLLL16} that integrates the attention distribution over the input words with the vocabulary distribution.

\subsection{Exponential Moving Average}

The exponential moving average (EMA) technique, also referred to as temporal averaging, was initially introduced to be used in optimization algorithms for better generalization performance and reducing noise from stochastic approximation in recent parameter estimates by averaging model parameters \cite{polyak1992acceleration,moulines2011non,DBLP:journals/corr/KingmaB14}.

In applying the technique, we maintain two sets of parameters:\ (1) training parameters $\theta$ that are trained as usual, and (2) evaluation parameters $\overline{\theta}$ that are an exponentially weighted moving average of the training parameters. The moving average is calculated using the following expression:
\begin{equation}
\overline{\theta} \xleftarrow{} \beta \times \overline{\theta}+(1-\beta) \times \theta
\end{equation}
where $\beta$ is the decay rate.
Previous work \cite{szegedy2016rethinking, merity2018regularizing,adhikari2019rethinking,liu2019learning} has used this technique for different tasks to produce more stable and accurate results. In Section \ref{results}, we show that using this simple technique considerably improves the performance of our model in both of the tasks.

\section{Experimental Setup}
In this section, we introduce the datasets first, then explain additional implementation details, and finally describe the evaluation metrics.

\subsection{Datasets}
We use the \textsc{WikiBio} dataset \cite{lebret-etal-2016-neural} for neural table-to-text generation. This dataset contains 728,321 articles from English Wikipedia and uses the first sentence of each article as the ground-truth description of the corresponding infobox. The dataset has been divided into training (80\%), validation (10\%), and test (10\%) sets.

For NQG, we use the SQuAD dataset v1.1 \cite{rajpurkar-etal-2016-squad} in our experiments, containing 536 Wikipedia articles with over 100K question-answer pairs. The test set of the original dataset is not publicly available.
Thus, \citet{du-etal-2017-learning} and \citet{zhou2017neural} re-divide available data into training, validation, and test sets, which we call \mbox{split-1} and \mbox{split-2}, respectively. In this paper, we conduct experiments and evaluate our model on both of the data splits.

\subsection{Implementation Details}
For the sake of reproducibility, we discuss implementation details for achieving the results shown in Tables \ref{table:2} and \ref{table:3}.
We train the model using cross-entropy loss and retain the model that works best on the validation set during training for both tasks.
We replace unknown tokens with a word from the input having the highest attention score. In addition, a decay rate of $0.9999$ is used for the exponential moving average in both of the tasks.

For the neural table-to-text generation task, we train the model up to 10 epochs with three different seeds and a batch size of 32.
We use a single-layer BiLSTM for the encoder and a single-layer LSTM for the decoder and set the dimension of the LSTM hidden states to 500.
Optimization is performed using the Adam optimizer with a learning rate of 0.0005 and gradient clipping when its norm exceeds 5. The word, field, and position embeddings are trainable and have a dimension of 400, 50, and 5, respectively. The maximum position number is set to 30. Any higher position number is therefore counted as 30. The most frequent 20,000 words and 1,480 fields in the training set are selected as word vocabulary and field vocabulary, respectively, for both the encoder and the decoder. Ultimately, we conduct greedy search to decode a description for a given input table.

For the NQG task, we use a two-layer BiLSTM for the encoder and a single-layer LSTM for the decoder. We set the dimension of the LSTM hidden states to 350 and 512 for split-1 and \mbox{split-2}, respectively.
Optimization is performed using the AdaGrad optimizer with a learning rate of 0.3 and gradient clipping when its norm exceeds 5.
The word embeddings are initialized with pre-trained 300-dimensional GloVe embeddings \cite{pennington2014glove}, which are frozen during training.
We train the model up to 20 epochs with five different seeds and a batch size of 50. We further employ dropout with a probability of 0.1 and 0.3 for data split-1 and split-2, respectively. Moreover, we use the vocabulary set released by \citet{song2018leveraging} for both the encoder and the decoder. During decoding, we perform beam search with a beam size of 20 and a length penalty weight of 1.75.

\subsection{Evaluation}

Following previous work, we use BLEU-4 \cite{papineni2002bleu}, METEOR \cite{banerjee2005meteor}, ROUGE-4, and ROUGE-L \cite{lin-2004-rouge} to evaluate the performance of our model. BLEU and METEOR were originally designed to evaluate machine translation systems, and ROUGE was designed to evaluate text summarization systems.

\begin{table*}[t!]
\centering
\resizebox{\textwidth}{!}{
\begin{tabular}{l|c|c|c|c|c|c}
\multirow{2}{*}{\textbf{Models}} & \multicolumn{3}{c|}{\textbf{Split-1}} & \multicolumn{3}{c}{\textbf{Split-2}}\\
\cline{2-7}                  
& \textbf{BLEU-4} & \textbf{METEOR} & \textbf{ROUGE-L} & \textbf{BLEU-4} & \textbf{METEOR} & \textbf{ROUGE-L}\\
\hline
\citet{Heilman:2011:AFQ:2520603}& - & - & - & 9.47 & 18.97 & 31.68 \\
\citet{du-etal-2017-learning} & 12.28 & 16.62 & 39.75 & - & - & -\\
\citet{zhou2017neural} & - & - & - & 13.29 & - & -\\
\citet{zhou2018sequential} & - & - & - & 13.02 & - & 44.0\\
\citet{ijcai2018-632} & - & - & - & 13.36 & 17.70 & 40.42\\
\citet{song2018leveraging} & 13.98 & 18.77 & 42.72 & 13.91 & - & -\\
\citet{zhao2018paragraph} &  15.32 & 19.29 & 43.91 & 15.82 & 19.67 & 44.24 \\
\citet{sun2018answer} &  - & - & - & 15.64 & - & - \\
\citet{kumar2018framework} & 16.17 & 19.85 & 43.90 & - & - & -\\
\citet{DBLP:conf/aaai/KimLSJ19} & 16.20 \textpm \ 0.32 &  19.92 \textpm \ 0.20 & 43.96 \textpm \ 0.25 & 16.17 \textpm \ 0.35& - & -\\
\citet{liu2019learning} &  - & - & - & \bfblue{17.55} & 21.24 & 44.53 \\
\hline  
Our Model & 14.81 \textpm \ 0.47 & 19.69 \textpm \ 0.24 & 43.01 \textpm \ 0.28 & 16.14 \textpm \ 0.25 & 20.44 \textpm \ 0.20 & 43.95 \textpm \ 0.26 \\
+ EMA & \bfblue{16.29 \textpm \ 0.04} & \bfblue{20.70 \textpm \ 0.08} & \bfblue{44.18 \textpm \ 0.15} & 17.47 \textpm \ 0.10 & \bfblue{21.37 \textpm \ 0.06} & \bfblue{45.18 \textpm \ 0.22} \\ 
\end{tabular}
}
\caption{Experimental results for NQG on the test sets.}
%\vspace{-.3cm}
\label{table:2}
\end{table*}

\begin{table}[t!]
\centering
\resizebox{0.47\textwidth}{!}{
\begin{tabular}{l|c|c}
\textbf{Models} &  \textbf{BLEU-4} & \textbf{ROUGE-4} \\
\hline
KN\footnotemark[1]& 2.21 & 0.38 \\
Template KN\footnotemark[7] & 19.80 & 10.70 \\
\citet{lebret-etal-2016-neural} & 34.70 \textpm \ 0.36 & 25.80 \textpm \ 0.36\\
\citet{bao2018table} & 40.26 & - \\
\citet{sha2018order} &  43.91 & 37.15\\
\citet{liu2018table} Orig. & 44.89 \textpm \ 0.33 &  41.21 \textpm \ 0.25 \\
\citet{liu2018table} Repl. & 44.45 \textpm \ 0.11 &  39.65 \textpm \ 0.10 \\
\citet{Liu_Luo_Xia_Ma_Chang_Sui_2019} & 45.14 \textpm \  0.34& 41.26 \textpm \ 0.37 \\
\hline
Our Model & 46.07 \textpm \ 0.17 & 41.53 \textpm \ 0.30 \\
+ EMA & \bfblue{46.76 \textpm \ 0.03} & \bfblue{43.54 \textpm \ 0.07} \\
\end{tabular}}
\caption{Experimental results for neural table-to-text generation on the test set. 
\footnotemark[1]KN is Kneser-Ney language model \cite{heafield2013scalable}. \footnotemark[7]Template KN is a KN language model over templates. Both models are proposed by \citet{lebret-etal-2016-neural} as baselines.
}
\label{table:3}
%\vspace{-.3cm}
\end{table}

\section{Results and Discussion} \label{results}
In this section, we present our experimental results for both neural table-to-text generation and NQG.
We report the mean and standard deviation of each metric across multiple seeds to ensure robustness against potentially spurious conclusions \cite{crane2018questionable}.
In Tables \ref{table:2} and \ref{table:3}, we compare previous work with our results for NQG and neural table-to-text generation, respectively. All results are copied from the original papers
except for \citet{liu2018table} in Table \ref{table:3}, where \textit{Repl.}\ refers to scores from experiments that we conducted using the source code released by the authors, and \textit{Orig.}\ refers to scores taken from the original paper.

It is noteworthy that a similar version of our model has served as a baseline in previous papers \cite{liu2018table,DBLP:conf/aaai/KimLSJ19,liu2019learning}.
However, the distinctions discussed in Section \ref{related_work}, especially the EMA technique, enable our model to achieve the state of the art in all cases but BLEU-4 on the SQuAD split-2, where our score is very competitive; furthermore, \citet{liu2019learning} only report results from a single trial.
Our results indicate that a basic seq2seq model is able to effectively learn the underlying distribution of both datasets. 

\section{Conclusions and Future Work}

In this paper, we question the necessity of complex neural architectures for text generation from structured data (neural table-to-text generation) and unstructured data (NQG).
We then propose a simple yet effective seq2seq model trained with the EMA technique. 
Empirically, our model achieves the state of the art in both of the tasks.
Our results highlight the importance of thoroughly exploring simple models before introducing complex neural architectures, so that we can properly attribute the source of performance gains.
As a potential direction for future work, it would be interesting to investigate the use of the EMA technique on transformer models as well and conduct similar studies to examine needless architectural complexity in other NLP tasks.

\section*{Acknowledgments}

This research was supported by the Natural Sciences and Engineering Research Council (NSERC) of Canada.

\bibliography{acl2020}

\begin{thebibliography}{39}
\expandafter\ifx\csname natexlab\endcsname\relax\def\natexlab#1{#1}\fi

\bibitem[{Adhikari et~al.(2019)Adhikari, Ram, Tang, and
  Lin}]{adhikari2019rethinking}
Ashutosh Adhikari, Achyudh Ram, Raphael Tang, and Jimmy Lin. 2019.
\newblock Rethinking complex neural network architectures for document
  classification.
\newblock In \emph{Proceedings of the 2019 Conference of the North American
  Chapter of the Association for Computational Linguistics: Human Language
  Technologies, Volume 1 (Long and Short Papers)}, pages 4046--4051.

\bibitem[{Bahdanau et~al.(2015)Bahdanau, Cho, and
  Bengio}]{DBLP:journals/corr/BahdanauCB14}
Dzmitry Bahdanau, Kyunghyun Cho, and Yoshua Bengio. 2015.
\newblock Neural machine translation by jointly learning to align and
  translate.
\newblock In \emph{Proceedings of the 3rd International Conference on Learning
  Representations}.

\bibitem[{Banerjee and Lavie(2005)}]{banerjee2005meteor}
Satanjeev Banerjee and Alon Lavie. 2005.
\newblock {METEOR}: An automatic metric for {MT} evaluation with improved
  correlation with human judgments.
\newblock In \emph{Proceedings of the {ACL} Workshop on Intrinsic and Extrinsic
  Evaluation Measures for Machine Translation and/or Summarization}, pages
  65--72.

\bibitem[{Bao et~al.(2018)Bao, Tang, Duan, Yan, Lv, Zhou, and
  Zhao}]{bao2018table}
Junwei Bao, Duyu Tang, Nan Duan, Zhao Yan, Yuanhua Lv, Ming Zhou, and Tiejun
  Zhao. 2018.
\newblock {Table-to-Text}: Describing table region with natural language.
\newblock In \emph{Proceedings of the Thirty-Second AAAI Conference on
  Artificial Intelligence}, pages 5020--5027.

\bibitem[{Crane(2018)}]{crane2018questionable}
Matt Crane. 2018.
\newblock Questionable answers in question answering research: Reproducibility
  and variability of published results.
\newblock \emph{Transactions of the Association of Computational Linguistics},
  6:241--252.

\bibitem[{Du et~al.(2017)Du, Shao, and Cardie}]{du-etal-2017-learning}
Xinya Du, Junru Shao, and Claire Cardie. 2017.
\newblock Learning to ask: Neural question generation for reading
  comprehension.
\newblock In \emph{Proceedings of the 55th Annual Meeting of the Association
  for Computational Linguistics (Volume 1: Long Papers)}, pages 1342--1352.

\bibitem[{Duan et~al.(2017)Duan, Tang, Chen, and Zhou}]{duan2017question}
Nan Duan, Duyu Tang, Peng Chen, and Ming Zhou. 2017.
\newblock Question generation for question answering.
\newblock In \emph{Proceedings of the 2017 Conference on Empirical Methods in
  Natural Language Processing}, pages 866--874.

\bibitem[{Gu et~al.(2016)Gu, Lu, Li, and Li}]{DBLP:conf/acl/GuLLL16}
Jiatao Gu, Zhengdong Lu, Hang Li, and Victor~O.K. Li. 2016.
\newblock Incorporating copying mechanism in sequence-to-sequence learning.
\newblock In \emph{Proceedings of the 54th Annual Meeting of the Association
  for Computational Linguistics (Volume 1: Long Papers)}, pages 1631--1640.

\bibitem[{Heafield et~al.(2013)Heafield, Pouzyrevsky, Clark, and
  Koehn}]{heafield2013scalable}
Kenneth Heafield, Ivan Pouzyrevsky, Jonathan~H. Clark, and Philipp Koehn. 2013.
\newblock Scalable modified {Kneser-Ney} language model estimation.
\newblock In \emph{Proceedings of the 51st Annual Meeting of the Association
  for Computational Linguistics (Volume 2: Short Papers)}, pages 690--696.

\bibitem[{Heilman(2011)}]{Heilman:2011:AFQ:2520603}
Michael Heilman. 2011.
\newblock \emph{Automatic Factual Question Generation from Text}.
\newblock Ph.D. thesis, Carnegie Mellon University.

\bibitem[{Heilman and Smith(2010)}]{heilman2010good}
Michael Heilman and Noah~A. Smith. 2010.
\newblock Good question! {Statistical} ranking for question generation.
\newblock In \emph{Human Language Technologies: The 2010 Annual Conference of
  the North American Chapter of the Association for Computational Linguistics},
  pages 609--617.

\bibitem[{Kim et~al.(2019)Kim, Lee, Shin, and Jung}]{DBLP:conf/aaai/KimLSJ19}
Yanghoon Kim, Hwanhee Lee, Joongbo Shin, and Kyomin Jung. 2019.
\newblock Improving neural question generation using answer separation.
\newblock In \emph{Proceedings of the Thirty-Third {AAAI} Conference on
  Artificial Intelligence}, pages 6602--6609.

\bibitem[{Kingma and Ba(2015)}]{DBLP:journals/corr/KingmaB14}
Diederik~P. Kingma and Jimmy Ba. 2015.
\newblock Adam: {A} method for stochastic optimization.
\newblock \emph{arXiv:1412.6980}.

\bibitem[{Kumar et~al.(2018)Kumar, Ramakrishnan, and Li}]{kumar2018framework}
Vishwajeet Kumar, Ganesh Ramakrishnan, and Yuan-Fang Li. 2018.
\newblock A framework for automatic question generation from text using deep
  reinforcement learning.
\newblock \emph{arXiv:1808.04961}.

\bibitem[{Lebret et~al.(2016)Lebret, Grangier, and
  Auli}]{lebret-etal-2016-neural}
R{\'e}mi Lebret, David Grangier, and Michael Auli. 2016.
\newblock Neural text generation from structured data with application to the
  biography domain.
\newblock In \emph{Proceedings of the 2016 Conference on Empirical Methods in
  Natural Language Processing}, pages 1203--1213.

\bibitem[{Lin(2004)}]{lin-2004-rouge}
Chin-Yew Lin. 2004.
\newblock {ROUGE}: A package for automatic evaluation of summaries.
\newblock In \emph{Text Summarization Branches Out}, pages 74--81, Barcelona,
  Spain.

\bibitem[{Lipton and Steinhardt(2018)}]{Lipton:1807.03341v2:2018}
Zachary~C. Lipton and Jacob Steinhardt. 2018.
\newblock Troubling trends in machine learning scholarship.
\newblock \emph{arXiv:1807.03341v2}.

\bibitem[{Liu et~al.(2019{\natexlab{a}})Liu, Zhao, Niu, Lai, He, Wei, and
  Xu}]{liu2019learning}
Bang Liu, Mingjun Zhao, Di~Niu, Kunfeng Lai, Yancheng He, Haojie Wei, and
  Yu~Xu. 2019{\natexlab{a}}.
\newblock Learning to generate questions by learning what not to generate.
\newblock \emph{arXiv:1902.10418}.

\bibitem[{Liu et~al.(2019{\natexlab{b}})Liu, Luo, Xia, Ma, Chang, and
  Sui}]{Liu_Luo_Xia_Ma_Chang_Sui_2019}
Tianyu Liu, Fuli Luo, Qiaolin Xia, Shuming Ma, Baobao Chang, and Zhifang Sui.
  2019{\natexlab{b}}.
\newblock Hierarchical encoder with auxiliary supervision for neural
  table-to-text generation: Learning better representation for tables.
\newblock In \emph{Proceedings of the Thirty-Third {AAAI} Conference on
  Artificial Intelligence}, pages 6786--6793.

\bibitem[{Liu et~al.(2018)Liu, Wang, Sha, Chang, and Sui}]{liu2018table}
Tianyu Liu, Kexiang Wang, Lei Sha, Baobao Chang, and Zhifang Sui. 2018.
\newblock Table-to-text generation by structure-aware seq2seq learning.
\newblock In \emph{Proceedings of the Thirty-Second AAAI Conference on
  Artificial Intelligence}, pages 4881--4888.

\bibitem[{Melis et~al.(2018)Melis, Dyer, and Blunsom}]{melis2018on}
Gábor Melis, Chris Dyer, and Phil Blunsom. 2018.
\newblock On the state of the art of evaluation in neural language models.
\newblock In \emph{Proceedings of the 6th International Conference on Learning
  Representations}.

\bibitem[{Merity et~al.(2018)Merity, Keskar, and
  Socher}]{merity2018regularizing}
Stephen Merity, Nitish~Shirish Keskar, and Richard Socher. 2018.
\newblock Regularizing and optimizing {LSTM} language models.
\newblock In \emph{Proceedings of the 6th International Conference on Learning
  Representations}.

\bibitem[{Mohammed et~al.(2018)Mohammed, Shi, and Lin}]{mohammed2017strong}
Salman Mohammed, Peng Shi, and Jimmy Lin. 2018.
\newblock Strong baselines for simple question answering over knowledge graphs
  with and without neural networks.
\newblock In \emph{Proceedings of the 2018 Conference of the North American
  Chapter of the Association for Computational Linguistics: Human Language
  Technologies, Volume 2 (Short Papers)}, pages 291--296.

\bibitem[{Mostafazadeh et~al.(2016)Mostafazadeh, Misra, Devlin, Mitchell, He,
  and Vanderwende}]{mostafazadeh-etal-2016-generating}
Nasrin Mostafazadeh, Ishan Misra, Jacob Devlin, Margaret Mitchell, Xiaodong He,
  and Lucy Vanderwende. 2016.
\newblock Generating natural questions about an image.
\newblock In \emph{Proceedings of the 54th Annual Meeting of the Association
  for Computational Linguistics (Volume 1: Long Papers)}, pages 1802--1813.

\bibitem[{Moulines and Bach(2011)}]{moulines2011non}
Eric Moulines and Francis~R. Bach. 2011.
\newblock Non-asymptotic analysis of stochastic approximation algorithms for
  machine learning.
\newblock In \emph{Advances in Neural Information Processing Systems}, pages
  451--459.

\bibitem[{Papineni et~al.(2002)Papineni, Roukos, Ward, and
  Zhu}]{papineni2002bleu}
Kishore Papineni, Salim Roukos, Todd Ward, and Wei-Jing Zhu. 2002.
\newblock {B}leu: a method for automatic evaluation of machine translation.
\newblock In \emph{Proceedings of the 40th Annual Meeting of the Association
  for Computational Linguistics}, pages 311--318.

\bibitem[{Pennington et~al.(2014)Pennington, Socher, and
  Manning}]{pennington2014glove}
Jeffrey Pennington, Richard Socher, and Christopher Manning. 2014.
\newblock {G}love: Global vectors for word representation.
\newblock In \emph{Proceedings of the 2014 Conference on Empirical Methods in
  Natural Language Processing ({EMNLP})}, pages 1532--1543.

\bibitem[{Polyak and Juditsky(1992)}]{polyak1992acceleration}
Boris~T. Polyak and Anatoli~B. Juditsky. 1992.
\newblock Acceleration of stochastic approximation by averaging.
\newblock \emph{SIAM Journal on Control and Optimization}, 30(4):838--855.

\bibitem[{Rajpurkar et~al.(2016)Rajpurkar, Zhang, Lopyrev, and
  Liang}]{rajpurkar-etal-2016-squad}
Pranav Rajpurkar, Jian Zhang, Konstantin Lopyrev, and Percy Liang. 2016.
\newblock {SQ}u{AD}: 100,000+ questions for machine comprehension of text.
\newblock In \emph{Proceedings of the 2016 Conference on Empirical Methods in
  Natural Language Processing}, pages 2383--2392.

\bibitem[{Sha et~al.(2018)Sha, Mou, Liu, Poupart, Li, Chang, and
  Sui}]{sha2018order}
Lei Sha, Lili Mou, Tianyu Liu, Pascal Poupart, Sujian Li, Baobao Chang, and
  Zhifang Sui. 2018.
\newblock Order-planning neural text generation from structured data.
\newblock In \emph{Proceedings of the Thirty-Second AAAI Conference on
  Artificial Intelligence}, pages 5414--5421.

\bibitem[{Song et~al.(2018)Song, Wang, Hamza, Zhang, and
  Gildea}]{song2018leveraging}
Linfeng Song, Zhiguo Wang, Wael Hamza, Yue Zhang, and Daniel Gildea. 2018.
\newblock Leveraging context information for natural question generation.
\newblock In \emph{Proceedings of the 2018 Conference of the North American
  Chapter of the Association for Computational Linguistics: Human Language
  Technologies, Volume 2 (Short Papers)}, pages 569--574.

\bibitem[{Sun et~al.(2018)Sun, Liu, Lyu, He, Ma, and Wang}]{sun2018answer}
Xingwu Sun, Jing Liu, Yajuan Lyu, Wei He, Yanjun Ma, and Shi Wang. 2018.
\newblock Answer-focused and position-aware neural question generation.
\newblock In \emph{Proceedings of the 2018 Conference on Empirical Methods in
  Natural Language Processing}, pages 3930--3939.

\bibitem[{Szegedy et~al.(2016)Szegedy, Vanhoucke, Ioffe, Shlens, and
  Wojna}]{szegedy2016rethinking}
Christian Szegedy, Vincent Vanhoucke, Sergey Ioffe, Jon Shlens, and Zbigniew
  Wojna. 2016.
\newblock Rethinking the inception architecture for computer vision.
\newblock In \emph{Proceedings of the IEEE Conference on Computer Vision and
  Pattern Recognition}, pages 2818--2826.

\bibitem[{Tang et~al.(2017)Tang, Duan, Qin, Yan, and Zhou}]{tang2017question}
Duyu Tang, Nan Duan, Tao Qin, Zhao Yan, and Ming Zhou. 2017.
\newblock Question answering and question generation as dual tasks.
\newblock \emph{arXiv:1706.02027}.

\bibitem[{Tang et~al.(2018)Tang, Duan, Yan, Zhang, Sun, Liu, Lv, and
  Zhou}]{tang2018learning}
Duyu Tang, Nan Duan, Zhao Yan, Zhirui Zhang, Yibo Sun, Shujie Liu, Yuanhua Lv,
  and Ming Zhou. 2018.
\newblock Learning to collaborate for question answering and asking.
\newblock In \emph{Proceedings of the 2018 Conference of the North American
  Chapter of the Association for Computational Linguistics: Human Language
  Technologies, Volume 1 (Long Papers)}, pages 1564--1574.

\bibitem[{Yao et~al.(2018)Yao, Zhang, Luo, Tao, and Wu}]{ijcai2018-632}
Kaichun Yao, Libo Zhang, Tiejian Luo, Lili Tao, and Yanjun Wu. 2018.
\newblock Teaching machines to ask questions.
\newblock In \emph{Proceedings of the Twenty-Seventh International Joint
  Conference on Artificial Intelligence, {IJCAI-18}}, pages 4546--4552.

\bibitem[{Zhao et~al.(2018)Zhao, Ni, Ding, and Ke}]{zhao2018paragraph}
Yao Zhao, Xiaochuan Ni, Yuanyuan Ding, and Qifa Ke. 2018.
\newblock Paragraph-level neural question generation with maxout pointer and
  gated self-attention networks.
\newblock In \emph{Proceedings of the 2018 Conference on Empirical Methods in
  Natural Language Processing}, pages 3901--3910.

\bibitem[{Zhou et~al.(2017)Zhou, Yang, Wei, Tan, Bao, and
  Zhou}]{zhou2017neural}
Qingyu Zhou, Nan Yang, Furu Wei, Chuanqi Tan, Hangbo Bao, and Ming Zhou. 2017.
\newblock Neural question generation from text: A preliminary study.
\newblock In \emph{National CCF Conference on Natural Language Processing and
  Chinese Computing}, pages 662--671. Springer.

\bibitem[{Zhou et~al.(2018)Zhou, Yang, Wei, and Zhou}]{zhou2018sequential}
Qingyu Zhou, Nan Yang, Furu Wei, and Ming Zhou. 2018.
\newblock Sequential copying networks.
\newblock In \emph{Proceedings of the Thirty-Second AAAI Conference on
  Artificial Intelligence}, pages 4987--4995.

\end{thebibliography}
\bibliographystyle{acl_natbib}

\end{document}